%% file: main.tex
\documentclass{article}

 \usepackage[preprint]{neurips_2026}

\usepackage[utf8]{inputenc} 
\usepackage[T1]{fontenc}    
\usepackage{hyperref}       
\usepackage{url}            
\usepackage{booktabs}       
\usepackage{amsfonts}       
\usepackage{nicefrac}       
\usepackage{microtype}      
\usepackage{xcolor}         
\usepackage{graphicx} 
\usepackage{amsmath}
\usepackage{float}
\usepackage{multirow}
\usepackage{subcaption}

\usepackage{amssymb}        
\usepackage{amsthm}         

\theoremstyle{definition}

\theoremstyle{remark}

\usepackage{array}
 
\title{When Language Overwrites Vision: Over-Alignment and Geometric Debiasing in Vision-Language Models}

\author{%
  Harshvardhan Saini\thanks{Equal contribution.} \\
  Indian Institute of Technology Dhanbad\\
  hs1062005@gmail.com\\
  \And
  Samyak Jha\footnotemark[1] \\
  Indian Institute of Technology Dhanbad\\
  samyakjha71@gmail.com\\
  \And
  Yiming Tang\thanks{Corresponding authors.} \\
  National University of Singapore\\
  yiming@nus.edu.sg \\
  \And
  Dianbo Liu\footnotemark[2] \\
  National University of Singapore\\
  dianbo@nus.edu.sg \\
}

\begin{document}
\maketitle

\input{sections/abstract}
\input{sections/introduction}
\input{sections/related_works}
\input{sections/analysis}
\input{sections/method}

\input{sections/results}
\input{sections/discussion}
\input{sections/limitations}

\newpage
\bibliography{main}
\bibliographystyle{plainnat}


\appendix
\newpage
\input{sections/appendix}


\end{document}

%% file: sections/abstract.tex
\begin{abstract}
    Vision-Language Models (VLMs) increasingly power high-stakes applications, from medical imaging to autonomous systems, yet they routinely hallucinate, confidently describing content not present in the input. We investigate the root causes of these failure modes with a mechanistic analysis focusing on the decoder-based VLMs. We trace these failure modes to a geometric over-alignment: to bridge the modality gap required by attention mechanisms, decoder-based VLMs over-align visual embeddings with the text manifold, injecting a statistical linguistic bias that systematically overshadows fine-grained visual evidence. While prior work either aggressively closes this gap or suppresses hallucinations through expensive black-box decoding strategies, none addresses the underlying geometric cause. We provide the first quantitative characterization of this over-alignment, demonstrating that linguistic bias concentrates in the top principal components of a universal, dataset-agnostic text subspace. Building on this insight, we propose two complementary remedies: a training-free inference strategy and a bias-aware fine-tuning paradigm, both of which explicitly project out this subspace from visual representations. Our methods significantly reduce hallucinations across POPE, CHAIR, and AMBER benchmarks, and improve CLAIR scores on long-form captioning tasks, with the training-free variant adding no computational overhead over the base model.
\end{abstract}

%% file: sections/introduction.tex
\section{Introduction}
Vision-Language Models (VLMs) have demonstrated remarkable performance across a wide variety of multi-modal tasks, powering applications from medical imaging to autonomous systems~\citep{bai2025qwen3,liu2023visual,zhu2025internvl3}. Standard decoder-based architectures typically consist of a vision encoder, a cross-modal projector, and a large language model (LLM) decoder. A central challenge reported in training and inference of these models is the \textit{modality gap}~\citep{liang2022mind,yu2026modality}: visual and linguistic representations naturally occupy disjoint manifolds within the shared latent space. This geometric misalignment has been closely linked to critical failure modes in VLMs, most notably \textit{object hallucination} — the confident generation of visual content that is simply not present in the image~\citep{rohrbach2018object}.

Prior work has explored the modality gap primarily in the context of CLIP-based dual-encoder architectures. \textit{Mind the Gap}~\citep{liang2022mind} provided an early characterization of this geometric separation, while subsequent approaches such as AlignVLM~\citep{masry2025alignvlm} and related projection techniques~\citep{yu2026modality} have attempted to close it by aggressively compressing visual representations into the text manifold. In decoder-based VLMs, bridging this gap is well motivated, since the attention mechanism requires both modalities to share a consistent representation space to compute ideal cross-modal similarity scores. However, whether closing this gap is the right strategy remains an open question~\citep{dhimoila2026cross}.

Parallel to efforts on the modality gap, a line of work has focused on mitigating hallucinations at the decoding stage, without directly addressing the modality gap. Methods such as VCD, SID, and CRoPS~\citep{leng2024mitigating,huo2024self,anand2026crops} employ contrastive or introspective decoding strategies to suppress hallucinated outputs. While effective to varying degrees, these approaches operate as computationally expensive black boxes and treat hallucination as a decoding problem rather than a representational one. ~\citet{dhimoila2026cross} employs Sparse Autoencoders (SAEs) to show that unimodal atoms explain the modality gap and act as modality-specific biases, but it requires computationally intensive dictionary learning and is validated on dual-encoder models.

Through a mechanistic analysis of typical decoder-based VLM architectures, we arrive at a key observation that challenges the prevailing assumption: a deliberate modality gap \textit{should} exist. Vision embeddings naturally inhabit a high-dimensional cone with broad angular spread, dense with fine-grained spatial and semantic structure, whereas the text manifold is comparatively compressed. Forcing these two spaces to fully overlap does not merely bridge a representational gap, it \textit{contaminates} the visual representations with spurious statistical regularities of language. To satisfy the alignment objective during pretraining, the model exploits a structural shortcut, injecting statistical textual patterns into the top principal components of the vision embedding space. These dominant directions carry no semantic signal about the actual image; they serve purely as a \textit{modality anchor}. As a result, the true visual signal is overshadowed by this textual bias, causing the model to hallucinate rather than attend to the actual image.

In this work, we present a mechanistic analysis and a geometric debiasing framework for decoder-based VLMs. We define a coordinate system for the text manifold via Principal Component Analysis (PCA) over pretraining captions, and provide the first quantitative characterization of over-alignment, showing that linguistic bias concentrates in the top PCs and actively degrades visual decodability across decoder layers. Building on this analysis, we propose two complementary remedies: a \textit{training-free inference strategy} that projects visual embeddings orthogonally to the top textual PCs at inference time with no additional computational overhead, and a \textit{bias-aware fine-tuning paradigm} that structurally removes the textual shortcut during training, allowing the cross-modal projector to natively encode visual evidence. Together, these methods significantly reduce hallucinations on POPE, CHAIR, and AMBER benchmarks, and improve CLAIR scores on long-form captioning tasks across multiple architectures.

Our main contributions are summarized as follows:
\begin{itemize}
    \item \textbf{Quantitative Characterization of Over-alignment:} We provide the first explicit analysis and quantification of the statistical textual bias induced by over-alignment in decoder-based VLMs, revealing how forced alignment suppresses fine-grained visual evidence.
    \item \textbf{Novel Geometric Framework:} We define a task-agnostic, text-based coordinate system to isolate and measure the penetration of linguistic bias within visual embeddings, identifying that this bias is heavily concentrated in the top principal components (PCs) of the latents.
    \item \textbf{Training-Free Debiasing Inference Strategy:} We introduce a principled, post-hoc debiasing method that systematically ablates top PCs to remove textual artifacts, significantly boosting visual grounding on diverse benchmarks without additional computation costs.
    \item \textbf{Bias-Aware Fine-Tuning Paradigm:} We propose a specialized training framework that explicitly deprives the network of statistical linguistic shortcuts during fine-tuning, forcing the model to natively ground its representations in visual evidence for trustworthy reasoning.
\end{itemize}
\begin{figure*}[t]
    \vspace{-2em}
    \includegraphics[width=\linewidth]{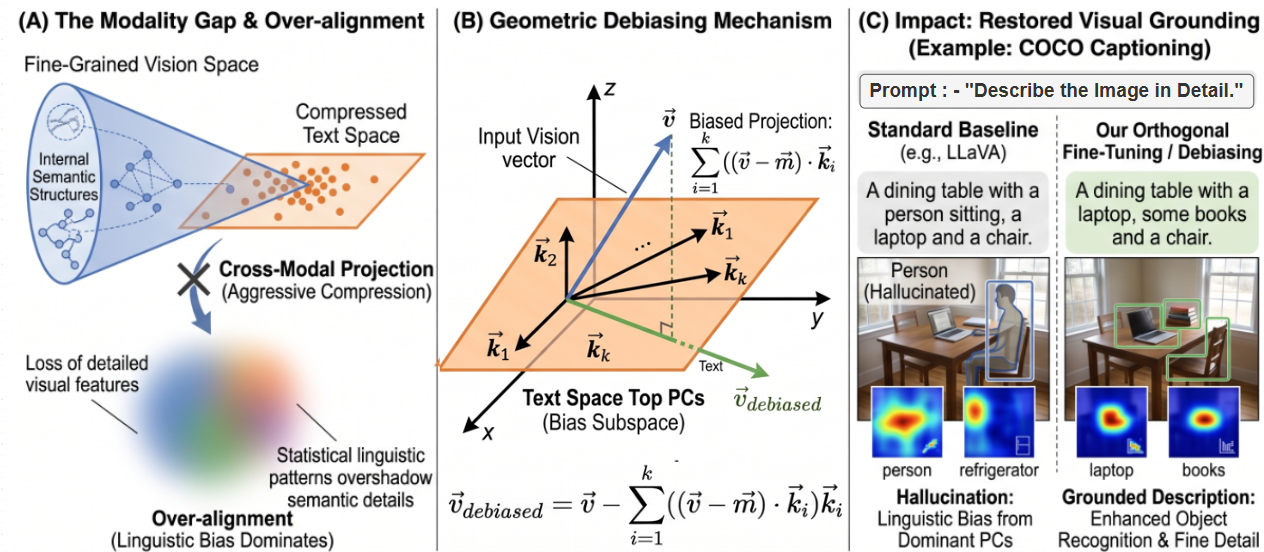}
    \caption{Overview of our geometric debiasing framework. (A) We identify that over-alignment with the text manifold suppresses visual details. (B) We propose a projection-based geometric intervention to isolate and remove statistical linguistic bias. (C) This method unmasks fine-grained visual evidence, directly reducing hallucinations and improving grounding accuracy.}
    \label{fig:hero_overview}
    \vspace{-1em}
\end{figure*}

%% file: sections/related_works.tex
\section{Related Works}

\subsection{Vision-Language Models and Modality Gap}
The rapid advancement of Large Language Models (LLMs) has enabled highly capable Vision-Language Models (VLMs), with diverse applications spanning visual question answering~\citep{agrawal2016vqavisualquestionanswering,tang2024demonstrationnotebookfindingsuited}, dataset analysis~\citep{luo2024llm}, and agentic systems~\citep{zou2026fmlbenchbenchmarkingmachinelearning}. These models typically couple a visual encoder with a pretrained language model backbone via a cross-modal projector~\citep{liu2023visual,bai2025qwen3,zhu2025internvl3}. As researchers investigate the training and inference pipelines of these VLMs~\citep{luo2023promptengineeringlensoptimal,wadekar2024evolutionmultimodalmodelarchitectures}, a central challenge emerges in the form of the Modality Gap~\citep{liang2022mind,yu2026modality}: the persistent geometric separation between vision and language embeddings within a shared latent space. This geometric misalignment has been shown to manifest in critical failure modes, most notably object hallucination~\citep{rohrbach2018object}, degraded visual grounding~\citep{li2023evaluating}, and an over-reliance on statistical textual priors at the expense of true visual evidence~\citep{leng2024mitigating}. While approaches such as AlignVLM~\citep{masry2025alignvlm} attempt to close this gap by enforcing stricter cross-modal alignment, we argue that such forced collapse constitutes over-alignment, introducing a statistical textual bias that actively degrades the semantic integrity of visual representations. There is literature on this statistical bias in papers such as VCD~\citep{leng2024mitigating}, SID~\citep{huo2024self} and CRoPS~\citep{anand2026crops},but they suppress the resulting hallucinations without addressing their geometric root cause in a black box manner; our work instead attempts to eliminate this bias directly at the representation level.

\subsection{Mechanistic Interpretability}

Mechanistic interpretability seeks to reverse-engineer the internal computations of neural networks by identifying the geometric and algebraic structure of their representations~\citep{sharkey2025openproblemsmechanisticinterpretability, zhao2026rep2textdecodingtextsingle}. A central organizing principle is the Linear Representation Hypothesis~\citep{elhage2022toy}, which posits that concepts are encoded as linear directions in activation space, motivating the use of Sparse Autoencoders (SAEs) to decompose representations into interpretable features~\citep{bricken2023monosemanticity, cunningham2023sparseautoencodershighlyinterpretable}. Within VLMs specifically, this lens has illuminated how visual features propagate through transformer layers~\citep{neo2024towards, liu2025visual}, characterized layer-wise token redundancy~\citep{chen2024image}, and enabled systematic analysis of generative model outputs~\citep{tang2025doesmodelfailautomatic, tang2025humanlikecontentanalysisgenerative}. Beyond analysis, sparse autoencoders have been extended to diverse applications including medical imaging~\citep{abdulaal2024xrayworth15features, tang2025cxrlaniclanguagegroundedinterpretableclassifier}, prompt engineering~\citep{saini2026bridgingmechanisticinterpretabilityprompt}, and model editing~\citep{wu2025axbenchsteeringllmssimple}. Complementing these empirical advances, rigorous theoretical frameworks have been proposed to characterize the foundations and limitations of sparse dictionary learning~\citep{tang2026unifiedtheorysparsedictionary, cui2026limitssparseautoencoderstheoretical}. Our work draws on this interpretability tradition to treat cross-modal textual bias as a linearly isolable subspace, enabling principled geometric debiasing of vision embeddings in decoder-based VLMs.

%% file: sections/analysis.tex
\section{Mechanistic Analysis}
Before introducing our mitigation strategies, we first empirically analyze the presence and impact of statistical textual bias within the vision embeddings. Standard VLM projection methods aggressively force visual representations into the text manifold. In this section, we investigate the mathematical necessity of this alignment, how we define the resulting geometric manifold, and how forced over-alignment alters the semantic properties of the vision tokens.

\subsection{The Mathematical Necessity of Alignment}
In decoder-based VLMs, bridging the modality gap is theoretically required to support the underlying attention mechanisms. Given a query matrix $Q$, key matrix $K$, and value matrix $V$, scaled dot-product attention is defined as:
$$Attention(Q,K,V)=\text{softmax}(\frac{QK^T}{\sqrt{d}})V$$
For the dot product $Q \cdot K^T$ to yield mathematically meaningful similarity scores between visual queries and textual keys (or vice versa), both modalities must operate within a shared, dimensionally consistent coordinate system. Thus, some degree of cross-modal alignment is unavoidable if we want to use Decoder-based LLM architecture.


\begin{figure*}[t]
    \vspace{-2em}
    \centering
    \begin{subfigure}[b]{0.48\textwidth}
        \centering
        \includegraphics[width=\textwidth]{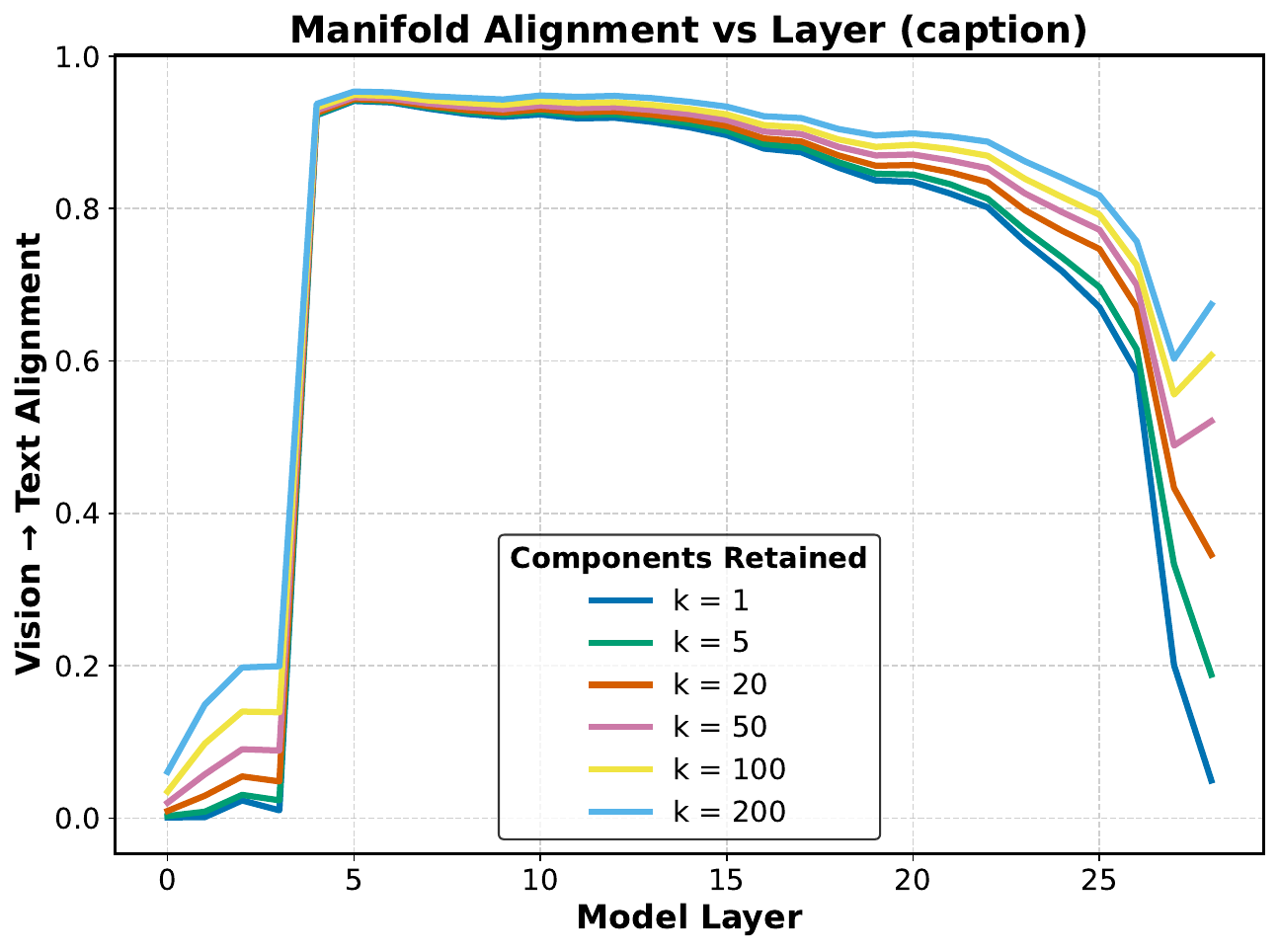}
        \caption{Full vision embeddings}
        \label{fig:alignment_full}
    \end{subfigure}
    \hfill
    \begin{subfigure}[b]{0.48\textwidth}
        \centering
        \includegraphics[width=\textwidth]{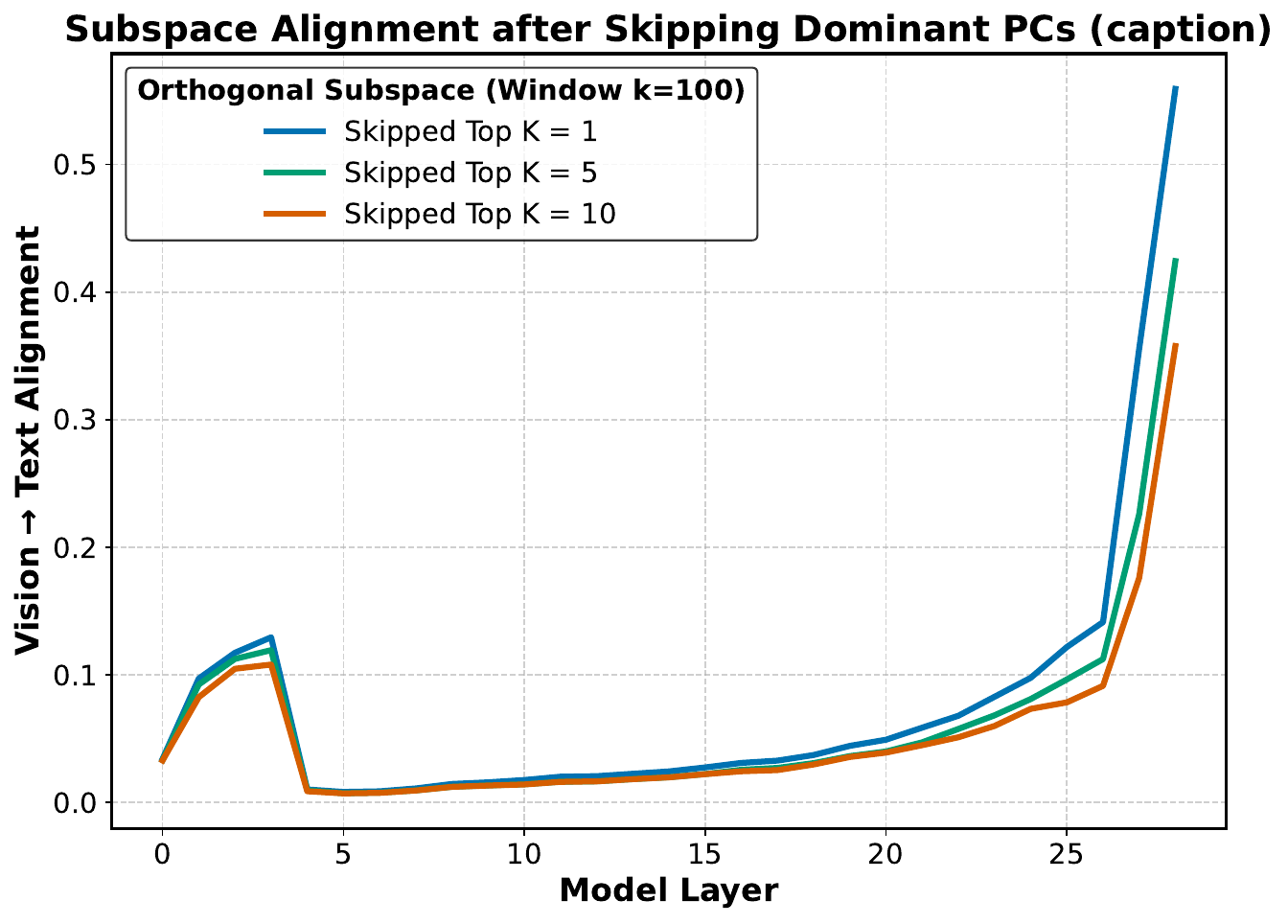}
        \caption{Top $k$ PCs removed}
        \label{fig:alignment_ablated}
    \end{subfigure}
    \caption{Layer-wise alignment scores of vision tokens onto the text manifold. (a) The alignment trajectory of the full vision embeddings demonstrates that the network aggressively forces the vision tokens to overlap with the textual space in the early to middle layers. (b) The alignment trajectory after removing the top $k$ principal components. The significant drop in alignment score confirms that the statistical textual bias is heavily concentrated in these top directions, leaving the lower PCs to occupy an orthogonal, modality-agnostic subspace.}
    \label{fig:layerwise_alignment}
    \vspace{-1em}
\end{figure*}

\subsection{Defining the Text Manifold and Universal Bias}
To investigate the geometry of this shared space, we must first rigorously define the text manifold. We find principal components (PCs) after isolating a representative set of captions directly from the pretraining dataset and taking vector embeddings of them. We compute the top Principal Components from these token embeddings, which allows us to represent the baseline linguistic distribution and cleanly isolate the core textual bias.
To verify the stability of this subspace, we compute and plot the Frobenius norm between text manifolds generated from multiple distinct text datasets in \ref{fig:pca_overlap}. Our analysis reveals that the top PCs form a consistently common manifold across all evaluated datasets. This suggests that standard pretraining injects a universal statistical linguistic bias rather than task-specific semantic structures.


\subsection{Layer-wise Alignment Trajectory}
To understand where and how the textual bias penetrates the vision embeddings, we analyze the layer-wise alignment of vision tokens within the LLM decoder. Building on our defined text manifold, we first center the vision embedding $\vec{v}^{(l)}$ at layer $l$ by subtracting the mean of the text manifold $\vec{m}$. We then project it onto the text-induced PCA basis $\{\vec{k}_{1},...,\vec{k}_{K}\}$ to obtain the text-aligned representation $\vec{z}^{(l)}$:
$$\vec{z}^{(l)} = \sum_{i} \left( (\vec{v}^{(l)} - \vec{m}) \cdot \vec{k}_i \right) \vec{k}_i$$
The alignment score is then defined as the ratio of the projected norm to the original centered norm:
$$Align(l)=\frac{1}{M}\sum_{j=1}^{M}\frac{||\vec{z}_{j}^{(l)}||_{2}}{||\vec{v}_{j}^{(l)}-\vec{m}||_{2}}$$
By analyzing the ratio of this projected norm to the original centered norm, we track the alignment trajectory. As shown in Figure \ref{fig:layerwise_alignment}a, the full vision embeddings exhibit an aggressively high alignment score in the early to middle layers, confirming that the network forces the vision tokens to heavily overlap with the text manifold. However, when we restrict this projection to manifolds defined without the top principal components (Figure \ref{fig:layerwise_alignment}b), the alignment scores drop significantly. This confirms our geometric hypothesis: the top PCs are hijacked to satisfy the alignment objective.

\subsection{Over-Alignment and the Overshadowing of Visual Details}
\begin{figure*}[t]
    \vspace{-2em}
    \centering

    \begin{subfigure}[b]{0.32\textwidth}
        \centering
        \includegraphics[width=\textwidth]{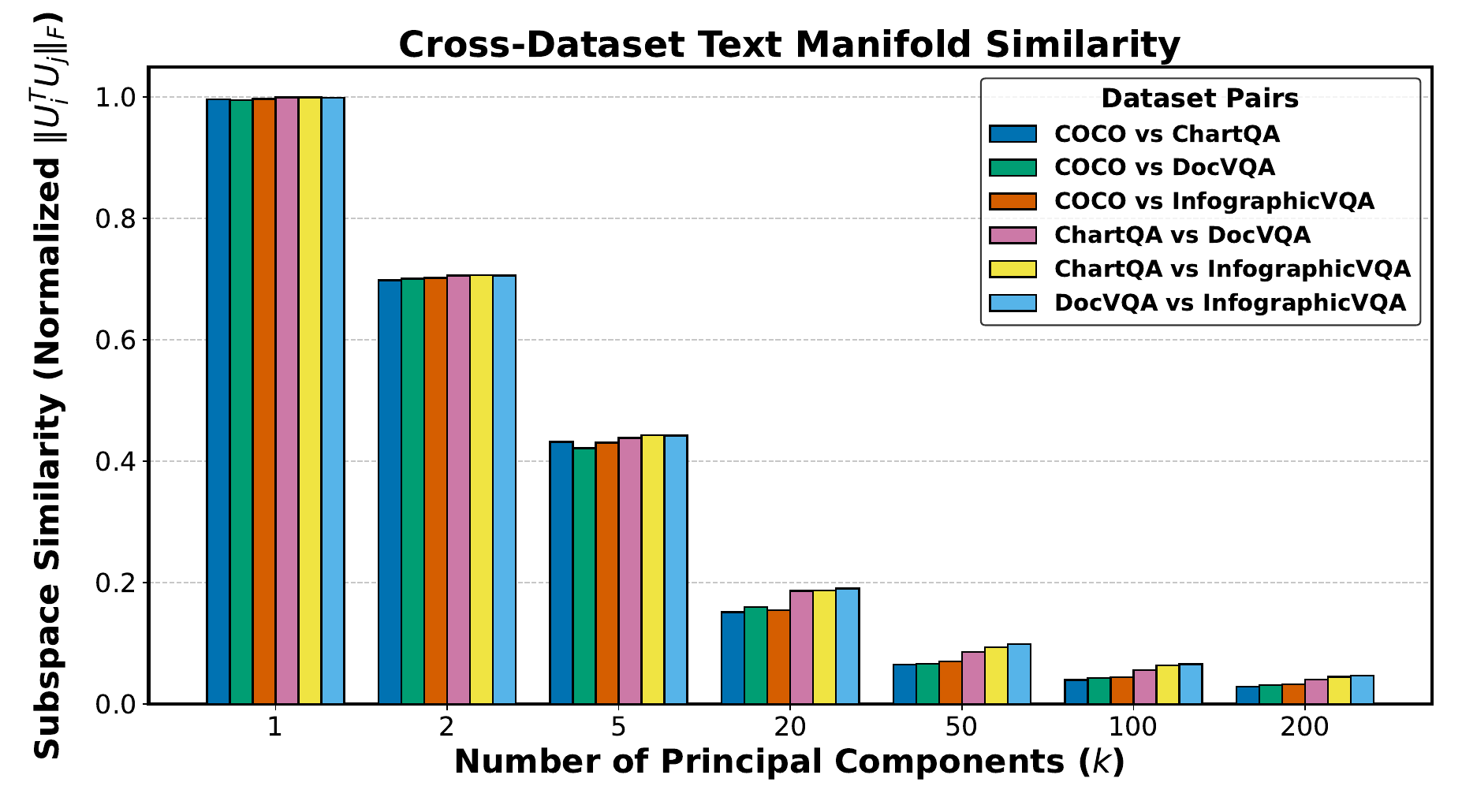}
        \caption{Cross-Dataset Text Manifold Similarity}
        \label{fig:pca_overlap}
    \end{subfigure}
    \hfill
    \begin{subfigure}[b]{0.32\textwidth}
        \centering
        \includegraphics[width=\textwidth]{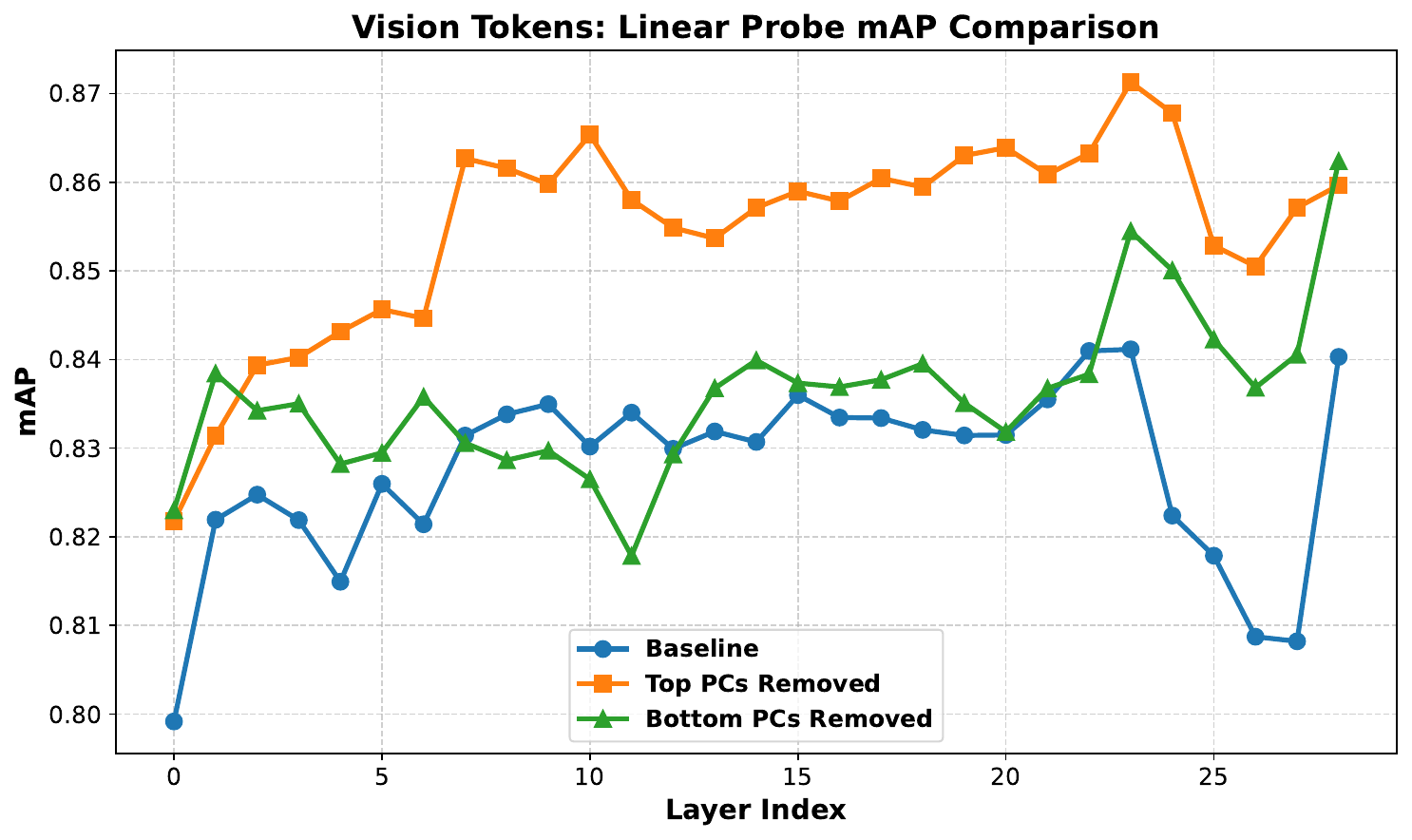}
        \caption{Linear Probe Performance (mAP)}
        \label{fig:linear_probe}
    \end{subfigure}
    \hfill
    \begin{subfigure}[b]{0.32\textwidth}
        \centering
        \includegraphics[width=\textwidth]{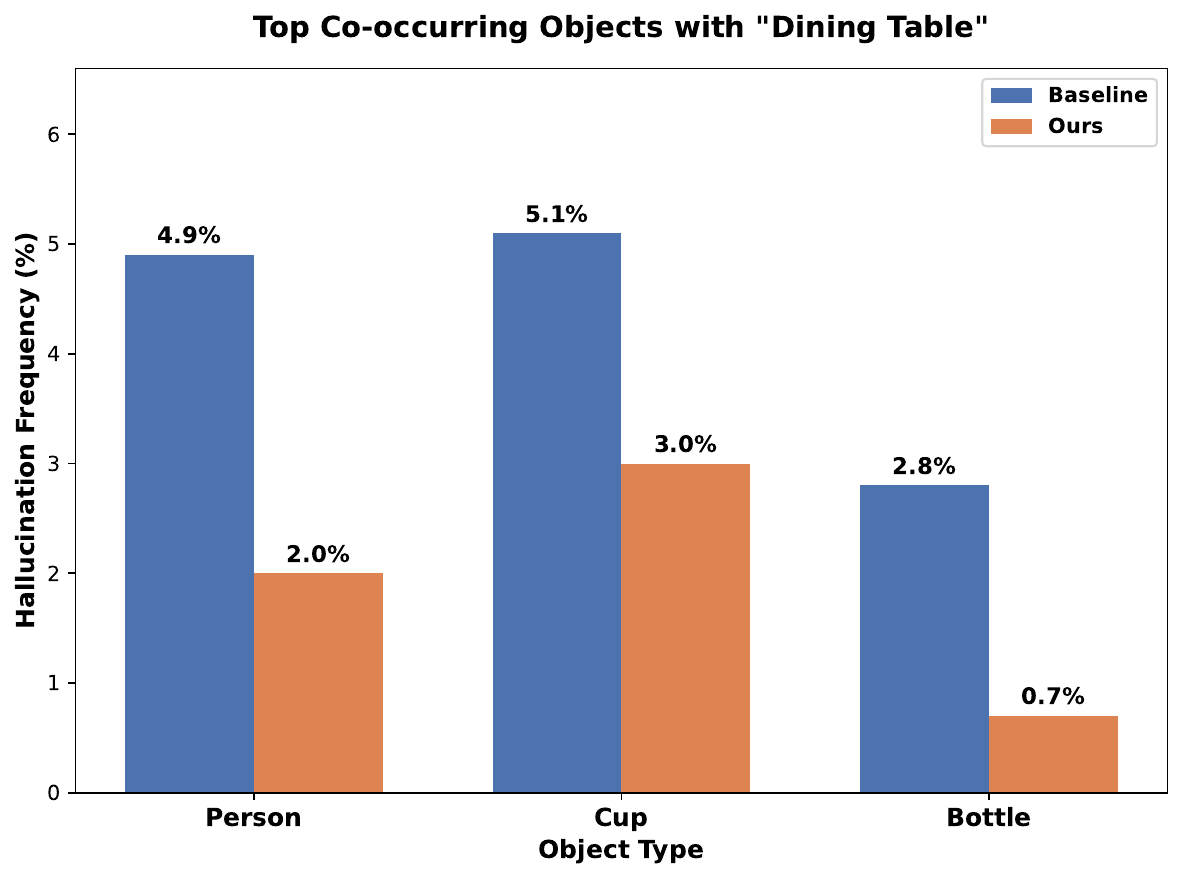}
        \caption{Frequency-Driven Hallucination}
        \label{fig:co_occurrence_hallucination}
    \end{subfigure}

    \caption{
    Geometric stability of textual bias and its impact on visual decodability.
    \textbf{(a)} Subspace similarity (Normalized $\|U_i^T U_j\|_F$) across diverse datasets. The near-perfect overlap in the top PCs confirms a universal linguistic bias, while lower PCs capture dataset-specific semantics.
    \textbf{(b)} Token-level linear probe performance across decoder layers. The baseline model (blue) degrades in middle-to-late layers due to structural text alignment. Projecting orthogonally to the top text PCs (orange) rescues and preserves fine-grained visual details.
    \textbf{(c)} Hallucination frequency of objects frequently co-occurring with ``dining table.'' The baseline model defaults to textual priors, while our method forces visual grounding and reduces hallucination rates by over 50\% on average.
    }

    \label{fig:geometry_and_probe}
    \vspace{-1.5em}
\end{figure*}

While Large Language Models rely on textual biases as essential priors to structure generation, enforcing these dataset biases onto vision embeddings overrides true visual evidence and acts as a structural distraction. We argue that the current pretraining paradigm suffers from over-alignment. To validate this, we conduct token-level linear probing for multi-label visual classification across all layers of the LLM decoder (Figure \ref{fig:geometry_and_probe}b). By extracting latent vision representations to train a regularized Ridge classifier on 80 MS-COCO object categories, we aggregate token-level logits via max-pooling to calculate image-level mean Average Precision (mAP). This setup measures precisely how linearly decodable the raw visual signal remains as it propagates through the network.

Throughout the decoder layers, the baseline model (blue) consistently yields the lowest decodability. When we perform a spectral intervention to project the vision tokens orthogonally to the top textual PCs (orange), linear decodability is significantly enhanced across the entire depth of the network. This continuous performance gap suggests that the top PCs act as a structural mask overriding the underlying visual data. Ultimately, these results demonstrate that standard projection forces visual details to compete with high-variance linguistic priors; by removing the top PCs, we strip away textual overshadowing and continuously unmask the localized visual details necessary for accurate semantic grounding.



\subsection{Geometrically Induced Hallucinations}
When dataset bias is left unchecked, it acts as a strong gravitational pull within the embedding space. Because the representations are skewed toward textual priors, the VLM undergoes visual overwrite, defaulting to high-probability text patterns instead of visually attending to the actual image. As illustrated in our frequency dependency plots, the model confidently predicts objects that frequently co-occur in text, ignoring their absence in reality. To empirically validate this visual overwrite, we analyze the hallucination frequency of objects that frequently co-occur in the pretraining corpus. We isolate image subsets containing a base object (e.g., ``dining table'') but strictly lacking common co-occurring objects (e.g., ``person,'' ``cup,'' or ``bottle''). As illustrated in Figure \ref{fig:co_occurrence_hallucination}, the baseline model frequently hallucinates these missing objects, driven by their statistical proximity in the text manifold as mentioned in paper~\citep{leng2024mitigating}. By applying our geometric debiasing strategy ablating the top principal components associated with the modality anchor we effectively sever this textual gravitational pull. Consequently, the hallucination rates drop substantially across all categories.


%% file: sections/method.tex
\section{Methodology}
Having established that over-alignment injects a statistical textual bias that acts as a structural distraction, we propose a geometric framework to extract and eliminate this bias. In this section, we formalize our text-induced coordinate system and detail our two proposed debiasing methods: a training-free inference strategy and a generalized VLM finetuning paradigm.

\subsection{Training-Free Debiasing Inference Strategy}
Our empirical analysis demonstrates that the top principal components of the vision embeddings encapsulate artifactual statistical bias. Building on the Linear Representation Hypothesis~\citep{elhage2022toy}, which posits that distinct concepts are encoded as linear directions within a representation space, we treat this textual bias as a linear subspace that can be cleanly isolated. To mitigate statistically induced hallucinations without requiring further training, we propose a post-hoc bias removal strategy. 

Given a vision embedding $\vec{v}$ and the mean of the text manifold $\vec{m}$, we first mean-center the vision embedding and compute its projection onto the top $k$ principal components of the text-aligned subspace. We then subtract this projection from the original embedding to obtain the debiased representation:
$$\vec{v}_{debiased}=\vec{v}-\sum_{i=1}^{k}((\vec{v}-\vec{m})\cdot\vec{k}_{i})\vec{k}_{i}$$

By projecting out the top PCs associated with the training captions, we explicitly remove the textual bias. This forces the VLM to rely on generalized semantic information, yielding embeddings that are crisp, semantic, and highly grounded.

\subsection{Bias-Aware Fine-Tuning Paradigm}
While our training-free strategy is effective at inference, we can further solidify this geometric debiasing by integrating it directly into the training loop. We propose a generalized VLM fine-tuning paradigm that explicitly prevents textual biases from abstracting visual embeddings during parameter updates. During the fine-tuning forward pass, we apply the identical top-$k$ PC component bias removal to the output of the cross-modal projector before it enters the LLM decoder. By structurally depriving the network of these top principal components during training, we force the VLM to rely on generalized semantic information rather than exploiting statistical textual shortcuts.
\label{fine-tuning}

%% file: sections/results.tex







\begin{table*}[t]
\centering
\scriptsize
\setlength{\tabcolsep}{3.2pt}
\renewcommand{\arraystretch}{0.95}

\resizebox{\textwidth}{!}{%
\begin{tabular}{llcccccccccc}
\toprule
\textbf{Model} & \textbf{Strategy} & \textbf{POPE} ($\uparrow$) & 
\multicolumn{2}{c}{\textbf{AMBER}} & 
\multicolumn{3}{c}{\textbf{CHAIR}} & 
\textbf{MME} ($\uparrow$) & 
\textbf{MMBench} ($\uparrow$) \\
\cmidrule(lr){4-5} \cmidrule(lr){6-8}
 &  &  & \textbf{CHAIR} ($\downarrow$) & \textbf{HAL} ($\downarrow$)
 & \textbf{CHAIR\_i} ($\downarrow$) & \textbf{CHAIR\_s} ($\downarrow$) & \textbf{Recall} ($\uparrow$)
 &  &  \\
\midrule

\multirow{6}{*}{Qwen2.5-VL-7B}
& Zero-shot 
& 87.3 
& 5.14 & 28.62
& 10.40 & 35.24 & 67.10 
& 2328.1 & 81.13 \\

& SID 
& 88.1 
& 4.58 & 25.90
& 8.87 & 30.60 & 58.50 
& 2342.8 & 82.15 \\

& VCD 
& 87.6 
& 4.72 & 26.35
& 8.75 & 33.10 & 60.95 
& 2335.7 & 82.55 \\

& VISTA 
& 88.0
& 4.46 & 25.79
& 8.95 & 33.00 & 60.49 
& 2338.1 & 82.70 \\

& DMAS 
& 88.3 
& 4.40 & 25.60
& 8.90 & 31.20 & 65.30 
& 2346.0 & 83.20 \\

& Ours 
& \textbf{88.4} 
& \textbf{4.23} & \textbf{25.07}
& \textbf{8.20} & \textbf{29.80} & \textbf{67.48} 
& \textbf{2355.2} & \textbf{83.33} \\

\midrule

\multirow{6}{*}{LLaVA-1.5-7B}
& Zero-shot 
& 82.8 
& 9.20 & 38.80
& 15.80 & 52.80 & 77.90 
& 1762.5 & 62.00 \\

& SID 
& 83.5 
& \textbf{6.24} & 33.73
& 14.20 & 50.60 & \textbf{77.94} 
& 1769.2 & 63.65 \\

& VCD 
& 83.3 
& 7.20 & 33.33
& 15.42 & 51.60 & 77.44 
& 1766.0 & 63.10 \\

& VISTA 
& 83.7 
& 6.90 & 31.94
& 13.28 & 46.40 & 75.74 
& 1760.0 & 62.70 \\

& DMAS 
& 83.6 
& 7.10 & 33.50
& 14.40 & 48.80 & 77.60 
& 1770.4 & 63.30 \\

& Ours 
& \textbf{83.8} 
& 6.74 & \textbf{29.87}
& \textbf{12.99} & \textbf{45.60} & 75.11 
& \textbf{1771.3} & \textbf{63.70} \\

\bottomrule
\end{tabular}
}

\caption{
Comparison of training-free hallucination mitigation methods across benchmarks.
Best values per model and metric are highlighted in bold.
}
\label{tab:hallucination_results}

\end{table*}

\begin{table}[t]
\centering
\footnotesize
\setlength{\tabcolsep}{5pt}
\renewcommand{\arraystretch}{1.05}

\caption{
Comparison of captioning performance on descriptive caption benchmarks using CLAIR score ($\uparrow$).
}

\begin{tabular}{llccc}
\toprule
\textbf{Dataset} & \textbf{Model} & \textbf{Zero-shot} & \textbf{Standard SFT} & \textbf{Ours} \\
\midrule

\multirow{2}{*}{VRSBench}
& Qwen2.5-VL-7B & 63.90 & 72.30 & \textbf{74.70} \\
& LLaVA-1.5-7B & 54.10 & 63.50 & \textbf{65.35} \\

\midrule

\multirow{2}{*}{TextCaps}
& Qwen2.5-VL-7B & 69.95 & 77.10 & \textbf{78.75} \\
& LLaVA-1.5-7B & 56.60 & 60.30 & \textbf{63.80} \\

\bottomrule
\end{tabular}

\label{tab:clair}
\end{table}
\section{Results}
In this section, we empirically validate our theoretical framework across two complementary settings: a training-free inference strategy and a fine-tuning paradigm. Together, these experiments demonstrate that geometric debiasing of vision embeddings reduces hallucinations and improves visual grounding regardless of whether bias removal is applied at inference time 
or baked into the training loop.

\subsection{Training-Free Bias Removal}
If the top principal components act purely as a linguistic bias to bridge the modality gap, ablating them during inference should reduce reliance on textual priors and force the model to ground its generations in visual evidence. We evaluate this post-hoc mitigation strategy on two prominent decoder-based architectures: LLaVA-1.5-7B and Qwen2.5-VL-7B. We report performance on standard benchmarks evaluating object hallucination and general capability: POPE~\citep{li2023evaluating}, AMBER~\citep{wang2023amber}, CHAIR~\citep{rohrbach2018object}, MME~\citep{fu2023mme}, and MMBench~\citep{liu2024mmbench}. We compare against established training-free baselines (VCD~\citep{leng2024mitigating}, SID~\citep{huo2024self}, VISTA~\citep{li2025hidden}, and DMAS~\citep{yin2026dynamic}).

As detailed in Table~\ref{tab:hallucination_results}, ablating the modality anchor consistently yields significant hallucination reductions while preserving core capabilities. On \textbf{Qwen2.5-VL-7B}, our method cuts AMBER CHAIR from 5.14 to 4.23 (-17.7\%) and HAL from 28.62 to 25.07 (-12.4\%). Similarly, CHAIR$_i$ drops from 10.40 to 8.20 (-21.2\%) and CHAIR$_s$ from 35.24 to 29.80 (-15.4\%). This hallucination reduction does not degrade visual grounding: POPE rises from 87.3 to 88.4, and Recall is preserved (67.10 $\rightarrow$ 67.48). General benchmarks also see slight improvements (MME: 2328.1 $\rightarrow$ 2355.2; MMBench: 81.13 $\rightarrow$ 83.33).

The improvements are even more pronounced on \textbf{LLaVA-1.5-7B}. Our approach cuts AMBER CHAIR from 9.20 to 6.74 (-26.7\%) and HAL from 38.80 to 29.87 (-23.0\%). Instance-level CHAIR$_i$ falls from 15.80 to 12.99 (-17.8\%), and sentence-level CHAIR$_s$ improves from 52.80 to 45.60 (-13.6\%). POPE improves from 82.8 to 83.8 while maintaining competitive Recall (77.90 $\rightarrow$ 75.11). Across both architectures, our lightweight, single-pass method matches or outperforms computationally heavy, multi-pass decoding baselines, proving that linguistic bias is a universal artifact of forced modality alignment.

\definecolor{box1}{HTML}{E41A1C}
\definecolor{box2}{HTML}{377EB8}
\definecolor{box3}{HTML}{4DAF4A}
\begin{figure*}[t]
    \centering


    \begin{subfigure}[t]{0.24\textwidth}
        \centering
        \includegraphics[
            width=\linewidth,
            height=3.3cm,
            keepaspectratio
        ]{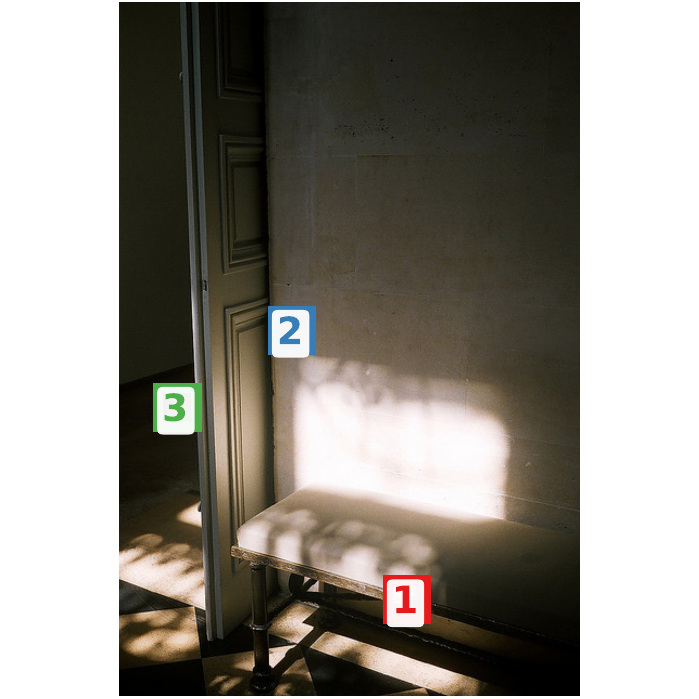}
        \caption{}
        \label{fig:qual_a}
    \end{subfigure}
    \hfill
    \begin{subfigure}[t]{0.24\textwidth}
        \centering
        \includegraphics[
            width=\linewidth,
            height=3.3cm,
            keepaspectratio
        ]{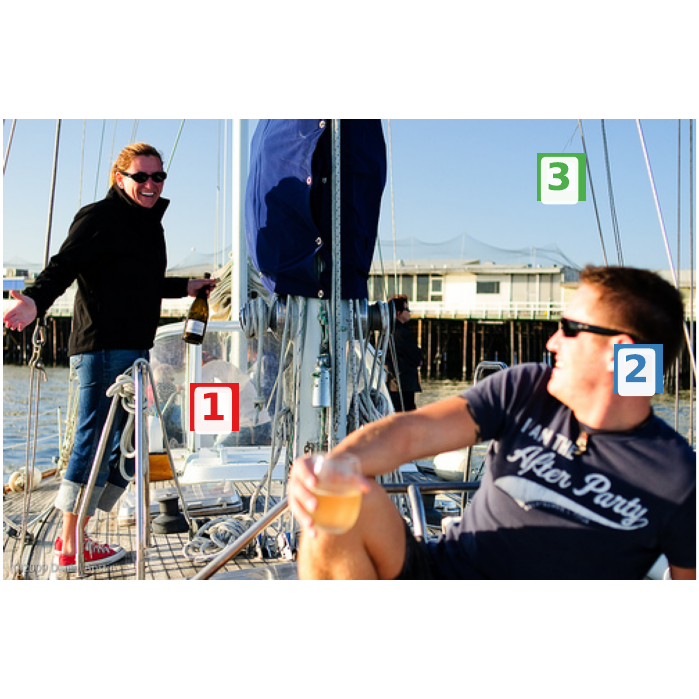}
        \caption{}
        \label{fig:qual_b}
    \end{subfigure}
    \hfill
    \begin{subfigure}[t]{0.24\textwidth}
        \centering
        \includegraphics[
            width=\linewidth,
            height=3.3cm,
            keepaspectratio
        ]{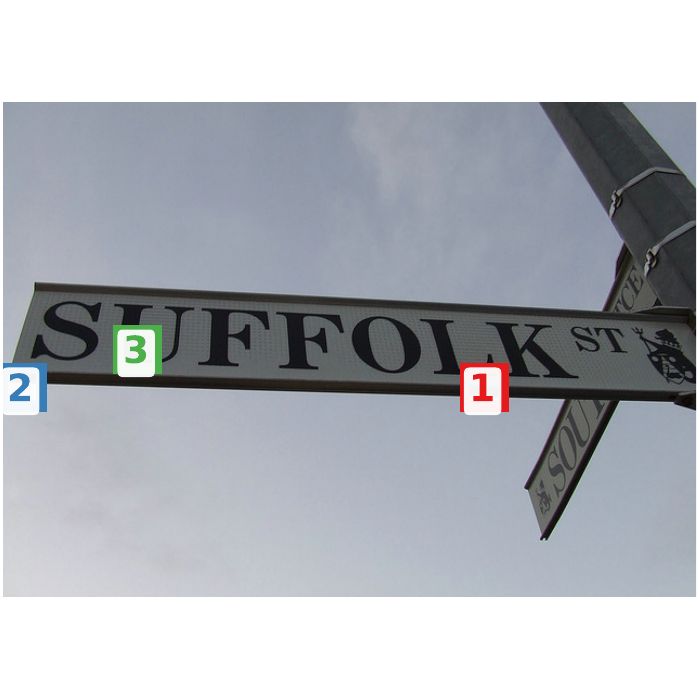}
        \caption{}
        \label{fig:qual_c}
    \end{subfigure}
    \hfill
    \begin{subfigure}[t]{0.24\textwidth}
        \centering
        \includegraphics[
            width=\linewidth,
            height=3.3cm,
            keepaspectratio
        ]{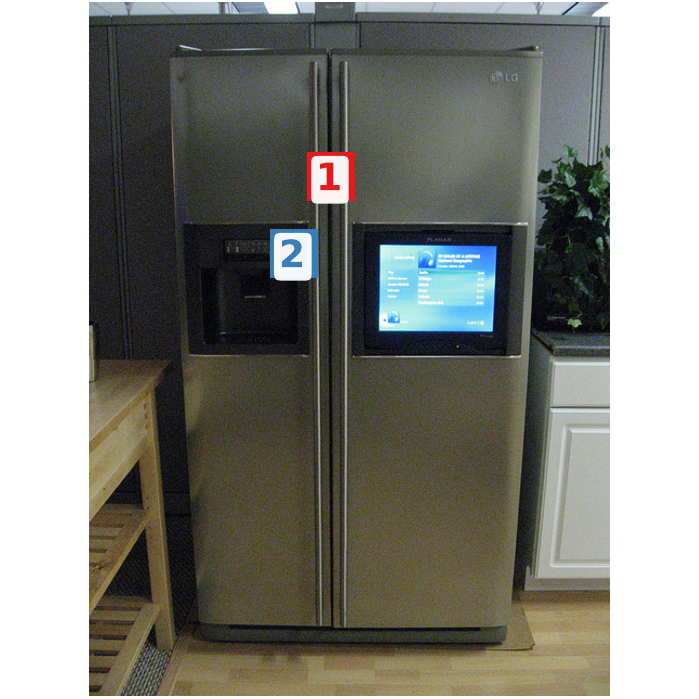}
        \caption{}
        \label{fig:qual_d}
    \end{subfigure}

    \vspace{1em}


    \resizebox{0.92\textwidth}{!}{
    \begin{tabular}{
        c
        c
        >{\ttfamily}l
        c
        >{\ttfamily}l
    }

        \toprule
        \textbf{Image} & \textbf{Patch} & \textbf{Top Token for Baseline (Over-Aligned)} & & \textbf{Top Token for Ours} \\
        \midrule

        \multirow{3}{*}{(a)}
        & \textcolor{box1}{\textbf{1}} & "," & $\rightarrow$ & "bench" \\
        & \textcolor{box2}{\textbf{2}} & "a" & $\rightarrow$ & "doorway" \\
        & \textcolor{box3}{\textbf{3}} & "a" & $\rightarrow$ & "open" \\
        \midrule

        \multirow{3}{*}{(b)}
        & \textcolor{box1}{\textbf{1}} & "," & $\rightarrow$ & "deck" \\
        & \textcolor{box2}{\textbf{2}} & "in" & $\rightarrow$ & "man" \\
        & \textcolor{box3}{\textbf{3}} & "," & $\rightarrow$ & "blue" \\
        \midrule

        \multirow{3}{*}{(c)}
        & \textcolor{box1}{\textbf{1}} & "<|object\_ref\_start|>" & $\rightarrow$ & "Street" \\
        & \textcolor{box2}{\textbf{2}} & "1" & $\rightarrow$ & "sign" \\
        & \textcolor{box3}{\textbf{3}} & "[empty]" & $\rightarrow$ & "U" \\
        \midrule

        \multirow{2}{*}{(d)}
        & \textcolor{box1}{\textbf{1}} & "by" & $\rightarrow$ & "refrigerator" \\
        & \textcolor{box2}{\textbf{2}} & "and" & $\rightarrow$ & "water" \\
        \bottomrule

    \end{tabular}
    }
    \vspace{0.5em}

    \caption{
    \textbf{Logit Lens Semantic Decoding.}
    By tracing the latent representations of specific image patches at later layers, we observe that the baseline VLM predominantly projects visual tokens into high-probability structural syntax (e.g., punctuation, prepositions, and articles). Removing the top principal components of the textual manifold unmasks the underlying orthogonal representation, recovering fine-grained visual semantics such as ``bench,'' ``refrigerator,'' and ``deck.''
    }
    \label{fig:logit_lens_qualitative}
    \vspace{-0.5cm}
\end{figure*}

\subsection{Bias Aware Fine-Tuning Results: Long-Form Captioning}
While the training-free strategy demonstrates the geometric debiasing principle at inference time, we further validate it by integrating bias removal directly into the fine-tuning loop (Section~\ref{fine-tuning}). To evaluate the fine-tuning paradigm, we move beyond short-answer 
hallucination benchmarks and specifically target \textbf{long-form descriptive captioning}, where the model must generate rich, open-ended descriptions of images. These tasks are particularly well-suited to exposing hallucination behavior at scale, since the model has greater freedom to inject statistically likely but visually absent objects.
For the purpose of finetuning we have kept the LLM frozen and finetuned only the projector and in only one stage for 2 epochs for each of the domains.
We evaluate on two domains: \textbf{VRSBench}~\citep{li2024vrsbench} and \textbf{TextCaps}~\citep{sidorov2020textcaps}. VRSBench tests captioning of remote sensing imagery, a domain with fine-grained spatial layouts that strongly punish models defaulting to common textual co-occurrence patterns. TextCaps requires the model to integrate text visible in images with visual context, demanding tight visual grounding. Both domains thus stress-test the core failure mode we identify: over-reliance on statistical textual priors at the expense of true visual evidence. Because traditional n-gram metrics such as BLEU or ROUGE measure surface-level lexical overlap 
and are poorly suited to evaluating long-form descriptive captions following ~\citet{li2024vrsbench}, we report the \textbf{CLAIR score}~\citep{chan2023clair}. CLAIR uses a language model to assess semantic fidelity between generated and reference captions, directly penalizing hallucinated content that does not correspond to the image while rewarding accurate visual grounding. This makes it 
a more faithful metric for our evaluation setting than n-gram alternatives.

As shown in Table~\ref{tab:clair}, our fine-tuning paradigm consistently outperforms both the zero-shot baseline and standard supervised fine-tuning (SFT) across both benchmarks and both architectures. On \textbf{VRSBench}, our method achieves a CLAIR score of 74.70 on Qwen2.5-VL-7B, surpassing the standard SFT baseline of 72.30 and the zero-shot baseline of 63.90. On LLaVA-1.5-7B, the improvement is similarly consistent, with our method reaching 65.35 versus standard SFT at 63.50 and zero-shot at 54.10. On \textbf{TextCaps}, gains are again uniform: Qwen2.5-VL-7B improves from 77.10 (standard SFT) to 78.75, and LLaVA-1.5-7B from 60.30 to 63.80, a particularly large jump that highlights the benefit of bias removal for a model that otherwise leans heavily on linguistic priors. Crucially, our method outperforms standard SFT despite sharing the same data and training budget, with the sole difference being the removal of the top principal components from vision embeddings during the forward pass. This confirms that the performance gains are attributable entirely to geometric debiasing rather than to additional data or compute. By depriving the network of its statistical textual shortcuts during training, we force it to natively ground its representations in visual evidence, yielding captions that are semantically richer, more accurate, and less prone to hallucination across diverse visual domains.

\subsection{Logit lens Analysis}

To explicitly trace the semantic information encoded within individual visual tokens prior to textual generation, we apply the Logit Lens technique to the latent representations at the later layer of the LLM decoder (Layer 26-28). By multiplying the hidden state of specific image patches directly with the unembedding matrix, we decode the underlying concepts dominating the token space.

As shown in Figure \ref{fig:logit_lens_qualitative}, the baseline representations are heavily distorted by the structural text components. This forces the model to decode visual patches as high-probability syntax markers (e.g., punctuation, articles, and prepositions like \texttt{","}, \texttt{"a"}, or \texttt{"in"}). When we geometrically ablate this universal textual bias, the underlying orthogonal representation is unmasked. The decoded tokens undergo a dramatic semantic shift, successfully bypassing the syntactic protocol to accurately reflect the fine-grained visual features of the corresponding patches, such as \texttt{"bench"}, \texttt{"refrigerator"}, and \texttt{"deck"}.

\subsection{Ablation Study}
\label{sec:ablation}

We perform two ablation studies on Qwen2.5-VL-7B using the CHAIR benchmark to analyze the sensitivity of our geometric intervention: 
(1) varying the number of removed top principal components ($K$), and 
(2) varying the decoder layer where the intervention is applied. 
We report CHAIR$_s$, CHAIR$_i$, and Recall to evaluate the trade-off between hallucination reduction and visual grounding.

\begin{figure*}[t]
    \centering

    \begin{subfigure}[t]{0.49\textwidth}
        \centering
        \includegraphics[width=\linewidth]{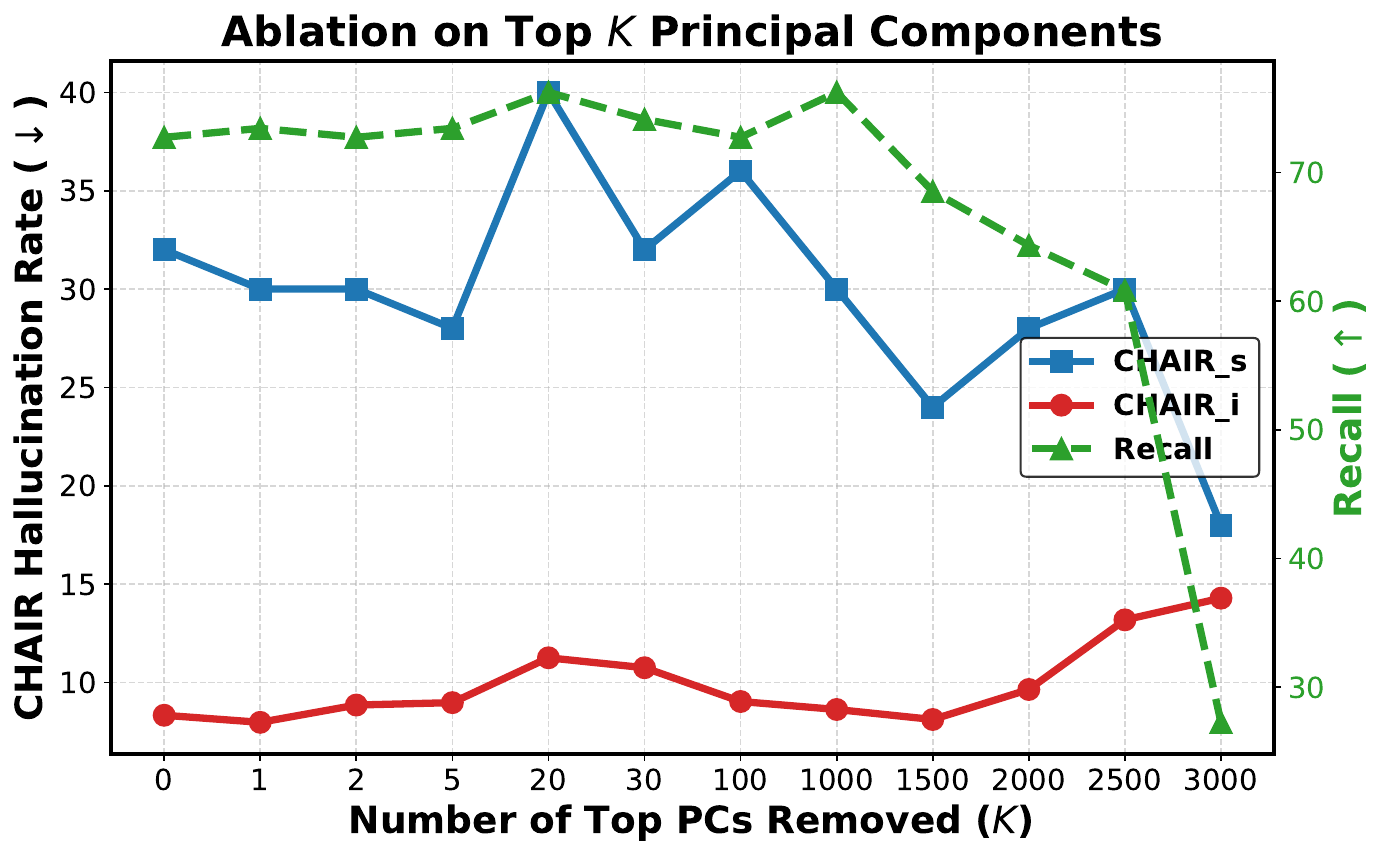}
        \label{fig:ablation_k}
        \caption{}
    \end{subfigure}
    \hfill
    \begin{subfigure}[t]{0.49\textwidth}
        \centering
        \includegraphics[width=\linewidth]{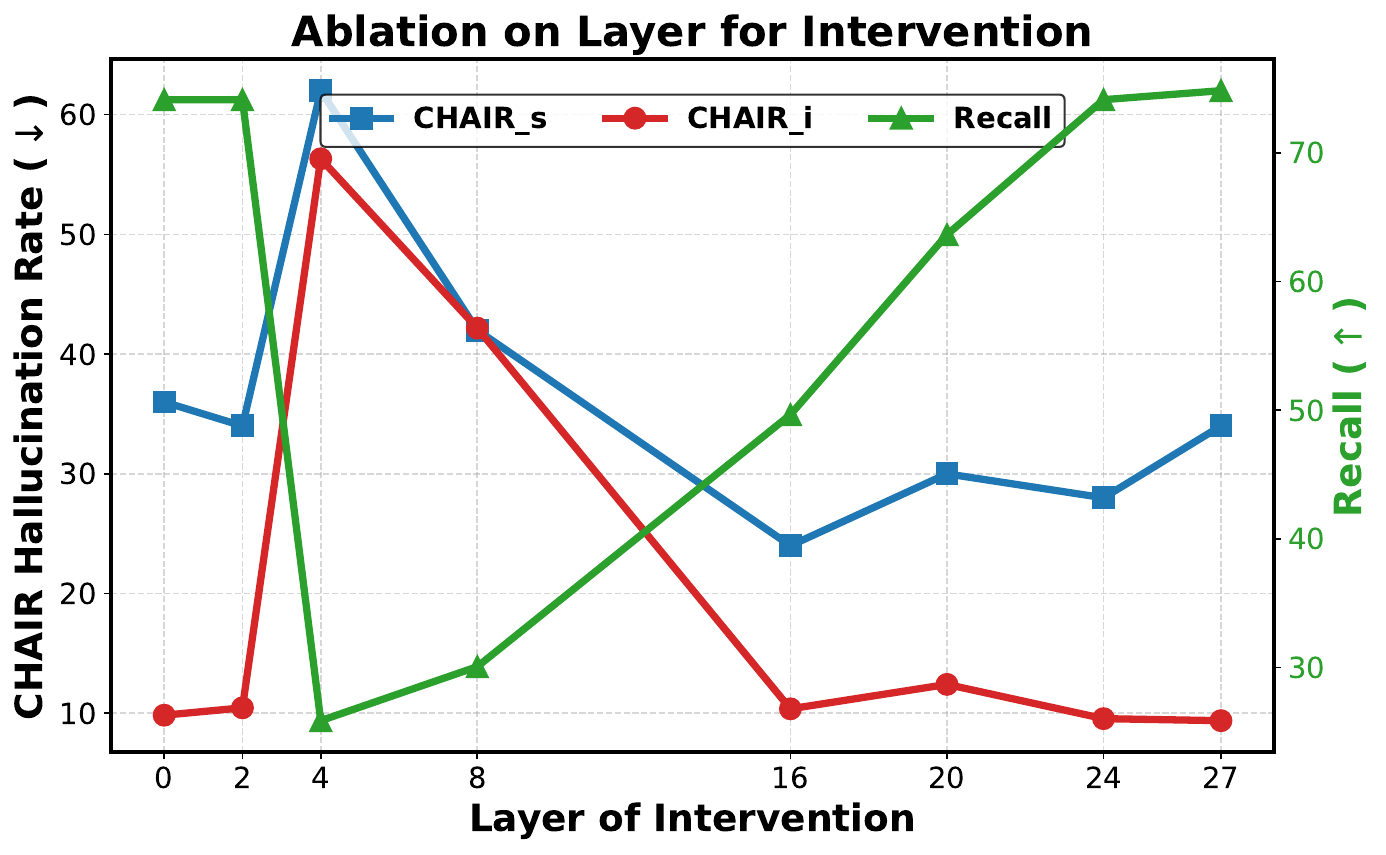}
        \label{fig:ablation_layer}
        \caption{}
    \end{subfigure}

    \caption{
    Ablation studies analyzing the sensitivity of geometric debiasing on the CHAIR benchmark.
    }
    
    \label{fig:ablation_study}
    \vspace{-0.5em}
\end{figure*}

%% file: sections/discussion.tex
\section{Discussion and Conclusion}
In this work, we mechanistically investigated over-alignment in decoder-based Vision-Language Models (VLMs). We demonstrated that forcing visual embeddings into a textual manifold to satisfy cross-modal attention injects a pervasive statistical linguistic bias, overshadowing fine-grained visual evidence and geometrically inducing object hallucinations. To address this, we introduced a geometric debiasing framework that quantifies and isolates this bias within the top principal components of the text manifold. Through both a training-free inference strategy and a bias-aware fine-tuning paradigm, our approach successfully severs the gravitational pull of linguistic priors. By unmasking the underlying orthogonal visual representations, our methods significantly reduce hallucinations and improve visual grounding across multiple benchmarks. Ultimately, our findings highlight the critical necessity of preserving the geometric integrity of visual representations to enable faithful, hallucination-free multimodal reasoning.

%% file: sections/limitations.tex
\section{Limitations}
While our geometric debiasing framework effectively mitigates hallucinations, we acknowledge fundamental limitations in our approach. First, we believe the current decoder-based LLM architecture is fundamentally flawed; the structural necessity of aggressively compressing a fine-grained visual space into a text manifold remains a bottleneck, suggesting that future architectures must likely rely on separate, modality-specific encoders to preserve semantic integrity. Second, our method depends on estimating a text manifold via Principal Component Analysis, which assumes statistical textual bias can be perfectly isolated into a linear subspace an assumption that may falter when the model encounters highly out-of-distribution domains.

%% file: sections/appendix.tex
\begin{appendix}
\section{Appendix}

\subsection{Implementation Details and Hyperparameters}
\label{app:hyperparameters}

For our fine-tuning experiments (Section \ref{fine-tuning}), we adopt a parameter-efficient approach designed to solely optimize the cross-modal projection space. Specifically, we freeze both the vision encoder and the large language model (LLM) backbone, limiting all gradient updates exclusively to the modality projector (the Merger in Qwen2.5-VL and the MLP adapter in LLaVA-1.5). During the forward pass, we apply our bias removal module to orthogonally decouple the top 2 principal components of the textual manifold, forcing the projector to learn fine-grained visual semantics.

The models are trained using the AdamW optimizer with a cosine learning rate scheduler. To maintain memory efficiency and training stability, we utilize DeepSpeed ZeRO-3 optimization and Mixed Precision (bfloat16) with TF32 enabled. Table \ref{tab:training_hyperparameters} details the exact hyperparameters used for fine-tuning both the Qwen2.5-VL-7B and LLaVA-1.5-7B architectures on the descriptive captioning datasets.

\begin{table}[h]
\centering
\small
\renewcommand{\arraystretch}{1.1}
\caption{Training hyperparameters for the geometric debiasing fine-tuning phase.}
\begin{tabular}{lcc}
\toprule
\textbf{Hyperparameter} & \textbf{Qwen2.5-VL-7B-Instruct} & \textbf{LLaVA-1.5-7B} \\
\midrule
Trainable Modules & Merger (Projector) & MLP Adapter (Projector) \\
Frozen Modules & Vision Encoder, LLM & Vision Encoder, LLM \\
Number of Top PCs Removed & 2 & 2 \\
Training Epochs & 2 & 2 \\
Global Batch Size & 126 & 48 \\
Optimizer & AdamW & AdamW \\
Learning Rate & 1e-3 & 2e-5 \\
Learning Rate Schedule & Cosine & Cosine \\
Warmup Ratio & 0.03 & 0.03 \\
Weight Decay & 0.1 & 0.0 \\
Precision & \texttt{bfloat16} (+ TF32) & \texttt{bfloat16} (+ TF32) \\
DeepSpeed Stage & ZeRO-3 (Offload) & ZeRO-3 \\
Gradient Checkpointing & True & True \\
\bottomrule
\end{tabular}
\label{tab:training_hyperparameters}
\end{table}

\subsection{Qualitative Analysis: Attention Score and Captions}
To further understand the effect of removing the modality anchor, we conduct a qualitative analysis using images from the COCO dataset. For each image, we prompt the model to list all objects that are clearly visible in the scene. We then compare the outputs produced by the baseline model and by our method after removing the top principal components associated with the modality anchor.

In addition to comparing the generated object lists, we analyze the cross-modal attention between generated textual tokens and visual patches. For each generated token, we aggregate the attention weights assigned to image patches and visualize them as spatial heatmaps over the input image. These heatmaps allow us to observe which regions of the image the model attends to while producing object descriptions.

\begin{figure}[htbp]
    \centering


    \begin{subfigure}[t]{0.48\textwidth}
        \centering
        \includegraphics[width=\textwidth]{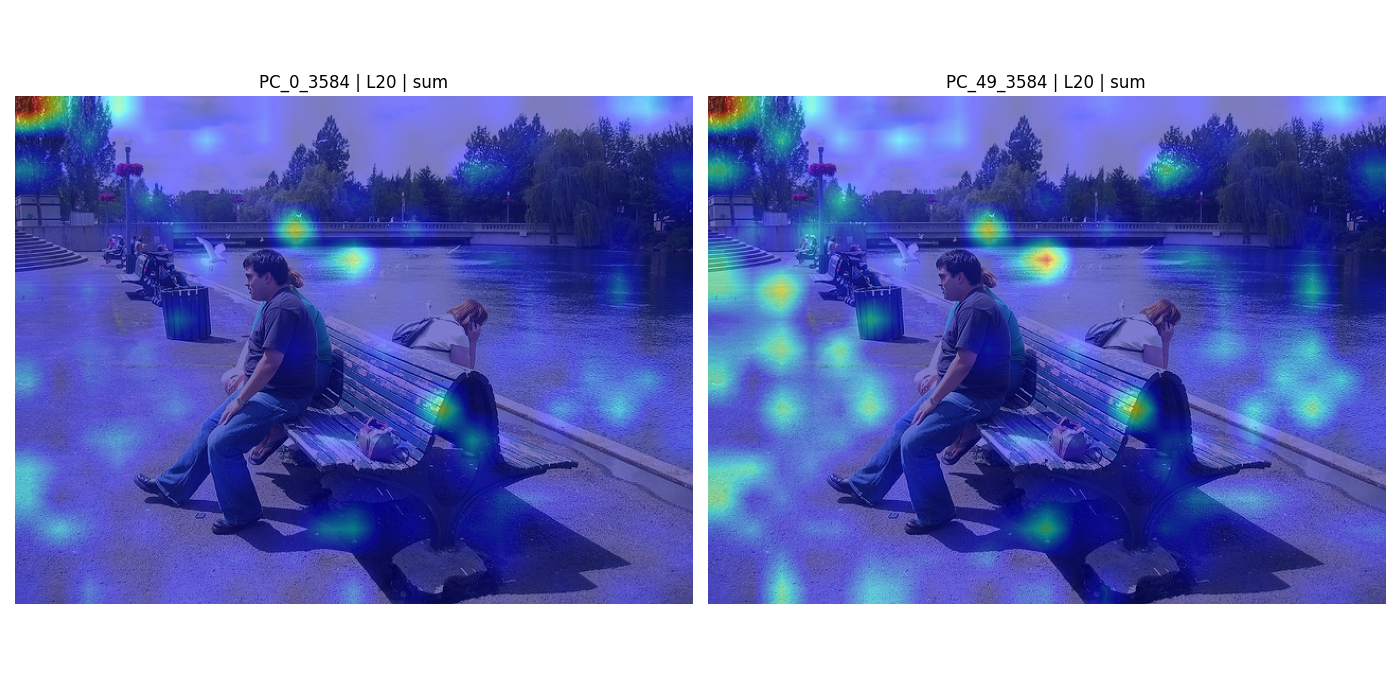}
        \caption{Park scene with improved grounding of distant objects.}
        \label{fig:qualitative_example1}
    \end{subfigure}
    \hfill
    \begin{subfigure}[t]{0.48\textwidth}
        \centering
        \includegraphics[width=\textwidth]{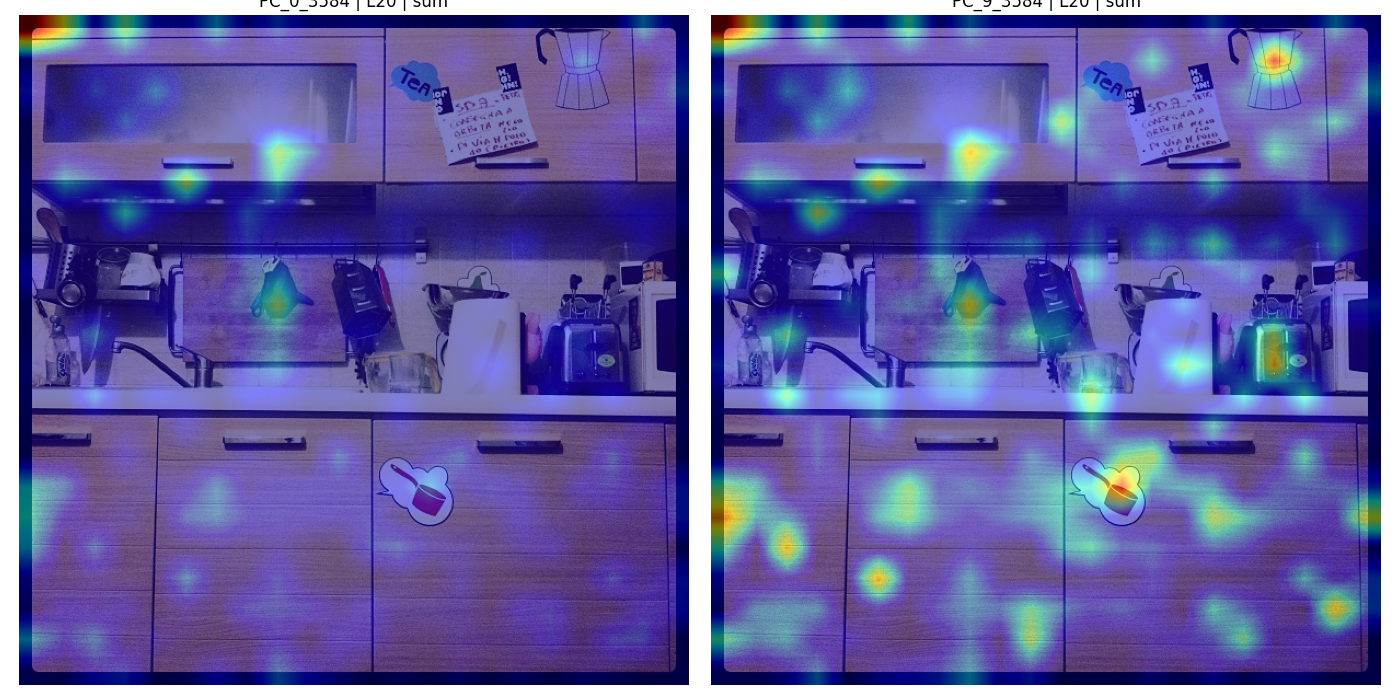}
        \caption{Kitchen scene with recovery of fine-grained objects.}
        \label{fig:qualitative_example3}
    \end{subfigure}

    \vspace{0.75em}


    \footnotesize
    \raggedright

    \textbf{(a) Baseline:}
    People sitting on a bench, bench with a bag on it, trash can, water body, trees and foliage, bridge in the background.

    \textbf{Ours:}
    People sitting on a bench, bench with a bag placed on it, body of water, trees and foliage in the background, bridge over the water, trash can near the bench, \textbf{birds flying above}, \textbf{staircase to the left}, \textbf{people walking in the distance}.

    \vspace{0.5em}

    \textbf{(b) Baseline:}
    Wooden cabinets, glass door cabinet, cutting board.

    \textbf{Ours:}
    Wooden cabinets, glass door cabinet, cutting board, \textbf{kitchen utensils}, \textbf{dish rack}, \textbf{coffee maker}, \textbf{tea kettle}, \textbf{stickers (speech bubbles)}, \textbf{notes}, \textbf{sink}, \textbf{dish soap bottle}, \textbf{sponge holder}.


    \caption{
    \textbf{Qualitative comparison of visual grounding via attention heatmaps.}
    The baseline model (left maps) suffers from attention smearing and focuses disproportionately on dominant foreground regions due to structural text bias. Removing the top principal components (right maps) sharpens the cross-modal attention, allowing the model to accurately detect and ground smaller, fine-grained objects that were previously ignored.
    }

    \label{fig:qualitative_comparison_combined}

\end{figure}

In our visualizations (Figure \ref{fig:qualitative_comparison_combined}), the left side of each panel corresponds to the baseline model, while the right side shows the model after removing the modality anchor. We observe that the baseline model often concentrates attention on a limited set of dominant foreground regions, overlooking smaller or less salient visual elements in the scene. In contrast, our method distributes attention across a broader set of semantically meaningful regions, enabling the model to detect additional objects that were previously ignored. In the captions accompanying each example, the \textbf{bolded phrases correspond to the objects} that are newly detected by our method but absent in the baseline output.

\end{appendix}